\documentclass[letterpaper, 10 pt, conference]{ieeeconf}  
\pdfoutput=1

\IEEEoverridecommandlockouts                              

\usepackage{graphics} 
\usepackage{subfigure}

\usepackage{epsfig} 
\usepackage{amsmath} 
\usepackage[T1]{fontenc}
\usepackage{bm}

\usepackage{booktabs}
\usepackage{multirow}
\usepackage{diagbox}
\usepackage{threeparttable}

\usepackage{tkz-euclide}
\usepackage{amsmath,amsfonts}
\usepackage{array}
\usepackage[caption=false,font=normalsize,labelfont=sf,textfont=sf]{subfig}
\usepackage{textcomp}
\usepackage{stfloats}
\usepackage{url}
\usepackage{verbatim}
\usepackage{graphicx}
\usepackage{cite}
\usepackage{times}
\usepackage{epsfig}
\usepackage{float}
\usepackage{wrapfig}
\usepackage{algorithm,algorithmicx,algpseudocode}
\usepackage{bm,xspace}
\usepackage{comment}
\usepackage{multirow}
\usepackage{balance}
\usepackage{booktabs}
\usepackage{etoolbox,siunitx}
\usepackage{calc}
\usepackage{pifont,hologo}
\usepackage{nicefrac}
\usepackage{makecell}
\usepackage{adjustbox}
\usepackage{bbding}
\usepackage{ragged2e}

\title{\LARGE \bf
NeRF-Based Transparent Object Grasping Enhanced by Shape Priors
}

\author{Yi Han$^{*}$, Zixin Lin$^{*}$, Dongjie Li, Lvping Chen, Yongliang Shi, and Gan Ma$^{\dag}$
\thanks{*Equal contribution, \dag  Corresponding author}
\thanks{Y. Han, Z. Lin, D. Li, L. Chen and G. Ma are with Sino-German College of Intelligent Manufacturing, Shenzhen Technology University, China. 
Y. Shi is with Tsinghua University, China.
Corresponding author: Gan Ma (magan@sztu.edu.cn)}}

\begin{document}

\maketitle
\thispagestyle{empty}
\pagestyle{empty}

\begin{abstract}
Transparent object grasping remains a persistent challenge in robotics, largely due to the difficulty of acquiring precise 3D information. Conventional optical 3D sensors struggle to capture transparent objects, and machine learning methods are often hindered by their reliance on high-quality datasets. 
Leveraging NeRF’s capability for continuous spatial opacity modeling, our proposed architecture integrates a NeRF-based approach for reconstructing the 3D information of transparent objects. Despite this, certain portions of the reconstructed 3D information may remain incomplete. To address these deficiencies, we introduce a shape-prior-driven completion mechanism, further refined by a geometric pose estimation method we have developed. This allows us to obtain a complete and reliable 3D information of transparent objects. Utilizing this refined data, we perform scene-level grasp prediction and deploy the results in real-world robotic systems. Experimental validation demonstrates the efficacy of our architecture, showcasing its capability to reliably capture 3D information of various transparent objects in cluttered scenes, and correspondingly, achieve high-quality, stable, and executable grasp predictions.

\end{abstract}

\section{Introduction}
In domestic robotics, grasping transparent objects is a critical challenge due to the difficulty in accurately capturing their 3D information.
The surfaces of transparent objects both reflect and refract light, which violates the Lambertian assumption on which most optical 3D sensors are based. Additionally, transparent objects lack prominent surface features, such as color and texture, resulting in a highly view-dependent appearance. These characteristics complicate the perception of transparent objects\cite{jiang2023robotic}.

Currently, many studies employ machine learning methods to estimate or complete the missing depth values of transparent objects \cite{sajjan2020clear,chen2022clearpose,wang2023mvtrans,fang2022transcg}. These approaches typically require large amounts of labeled data containing transparent objects, which implies longer data collection times. Moreover, obtaining accurate depth labels for transparent objects is inherently challenging. In addition, the lack of out-of-distribution generalization is a common obstacle for deep learning methods.

NeRF-based methods\cite{mildenhall2021nerf} provide a new approach to address the challenge of acquiring 3D information for transparent objects. As a technique that uses neural implicit fields to perform 3D scene reconstruction from multiple views, NeRF optimization is typically scene-specific and does not require learning, thereby avoiding the training and generalization issues associated with conventional deep learning. However, due to the optical principles which NeRF follows, such methods are generally sensitive to factors like ambient lighting conditions, the quality and number of views, and the optical properties of objects.
Consequently, these methods do not always ensure flawless visual reconstructions, with potential for incompleteness. Recent studies have underscored the utility of shape-prior-based completion techniques in restoring missing 3D information\cite{shi2024city}.


Our task is framed as the reliable grasping of transparent objects within a desktop scene. To achieve this, the proposed architecture is structured into the following stages: (1) 3D panoramic reconstruction of the desktop scene; (2) segmentation and pose estimation of objects within the scene; (3) shape completion for objects exhibiting incomplete point clouds; and (4) grasp prediction for transparent objects in the reconstructed scene. 
The panoramic reconstruction of the desktop scene is conducted using a NeRF-based neural network\cite{muller2022instant}, leveraging NeRF's ability to continuously model spatial opacity to capture the 3D information of transparent objects. We utilize the spatial distribution characteristics of the scene for object segmentation, and propose a geometry-driven pose estimation method that leverages specific object features, such as non-rotational symmetry, to support normalization during the shape completion process. For shape completion, we introduce a pre-trained auto-decoder informed by shape priors\cite{park2019deepsdf}, enabling the network to learn the geometric properties of similar objects and subsequently apply this knowledge to reconstruct incomplete transparent objects. Grasp predictions are generated using a model derived from GraspNet-1billion\cite{fang2020graspnet}, a widely adopted framework for grasp prediction in scene point clouds, and these predictions are further validated in real-world robotic systems.

To summarize, our contributions are as follows:
\begin{enumerate}
    \item[$\bullet$] A robust vision-based architecture for transparent object grasping, enhanced by shape priors and neural 3D scene reconstruction;  
    \item[$\bullet$] A dense surface shape completion method for sparse object point clouds, supported by a non-learning, geometry-driven pose estimation and normalization approach for non-revolute symmetric objects;
    \item[$\bullet$] Evaluation of the proposed architecture through grasping several transparent objects in cluttered scenes on a real robotic system, validating its effectiveness.
     
\end{enumerate}

\section{Relative Works}

\subsection{Transparent Object 3D Reconstruction}

Reconstructing transparent objects in 3D is challenging due to their optical properties, such as refraction and reflection. Various machine learning methods have been explored to estimate and complete missing depth information. Some approaches combine RGB-D data for depth estimation and shape reconstruction, but they face difficulties handling the complexity of transparent materials\cite{sajjan2020clear}. Other techniques use zero-shot transfer learning to enhance depth estimation, though optical distortions remain an issue\cite{bhat2023zoedepth}. Domain randomization has also been employed to simulate noisy conditions, improving depth perception but often struggling with reflections\cite{dai2022domain}.
Recent advances in NeRF (Neural Radiance Fields) show promise for transparent object reconstruction by learning volumetric representations. NeRF-W\cite{martin2021nerf} improves NeRF’s applicability to diverse scenes, while PlenOctrees\cite{yu2021plenoctrees} integrates octree structures for more efficient rendering. However, these methods require significant computational resources and long rendering times, limiting their practicality for real-time use\cite{barron2021mip, barron2022mip, barron2023zip, ichnowski2021dex}.
To address this, we employ instant-ngp\cite{muller2022instant}, a NeRF-based approach that offers faster processing with high reconstruction quality. Utilizing multi-resolution hash grids, instant-ngp is ideal for real-time scene reconstruction tasks, such as robotic manipulation in cluttered environments, balancing efficiency and quality for transparent object reconstruction.
 
\subsection{Shape Completion}

Shape completion from sparse or partial point clouds is a challenging problem, and various approaches have been proposed to tackle this issue, each with its limitations. Multi-view depth-based approaches \cite{lin2018learning,arsalan2017synthesizing,smith2018multi} rely on 2.5D depth maps to infer geometry but are often constrained by the availability of depth data and struggle with sparse inputs.
Voxel-based methods \cite{maturana2015voxnet,choy20163d} divide the 3D space into a grid of voxels and predict the occupancy of each voxel. Although straightforward, these methods suffer from high memory usage and tend to lose finer details at lower resolutions. Methods like ConvONet and Occupancy Networks \cite{mescheder2019occupancy,peng2020convolutional} also face challenges in recovering high-resolution details due to their reliance on discretized space representations, making them less suitable for handling complex geometries.
Patch-based approaches, such as AtlasNet \cite{goueix2018atlasnet}, fit a collection of surface patches to the point cloud. However, these methods are limited by their dependence on pre-defined patches, which makes them less effective for handling complex or irregular shapes.
To address the challenges of incomplete and sparse point clouds, we employ a DeepSDF-based approach\cite{park2019deepsdf} for shape completion. This method is particularly advantageous in generating smooth and detailed surfaces, even from sparse observations, and handles a broader range of shapes compared to template-based methods. Moreover, DeepSDF excels at representing continuous surface geometry and remains robust in reconstructing high-resolution details.

\subsection{Grasp Prediction}

Vision-based grasp prediction methods can be broadly categorized into two types: 2D planar grasp and 6-DoF grasp, both of which heavily rely on deep learning techniques.
In the realm of 2D planar grasping, methods detect graspable rectangular regions of the target object based on RGB-D image inputs \cite{lenz2015deep, redmon2015real, vohra2019real, kumra2017robotic, zhang2019roi, pharswan2020domain}. Mahler et al. \cite{mahler2017dex} introduced a grasp quality CNN trained on a substantial dataset to rank and identify the optimal grasp predictions. However, due to the limited degrees of freedom in 2D planar grasps, some approaches \cite{deng2020self} have focused on 6-DoF grasp prediction, projecting these predictions into the scene.
In our approach, we employ the GraspNet-1billion model \cite{fang2020graspnet} because it is well-suited for generating 6-DoF grasp predictions in cluttered desktop scenes.

\section{Method}

\begin{figure*}[!h]
    \centering
    \includegraphics[width=15cm]{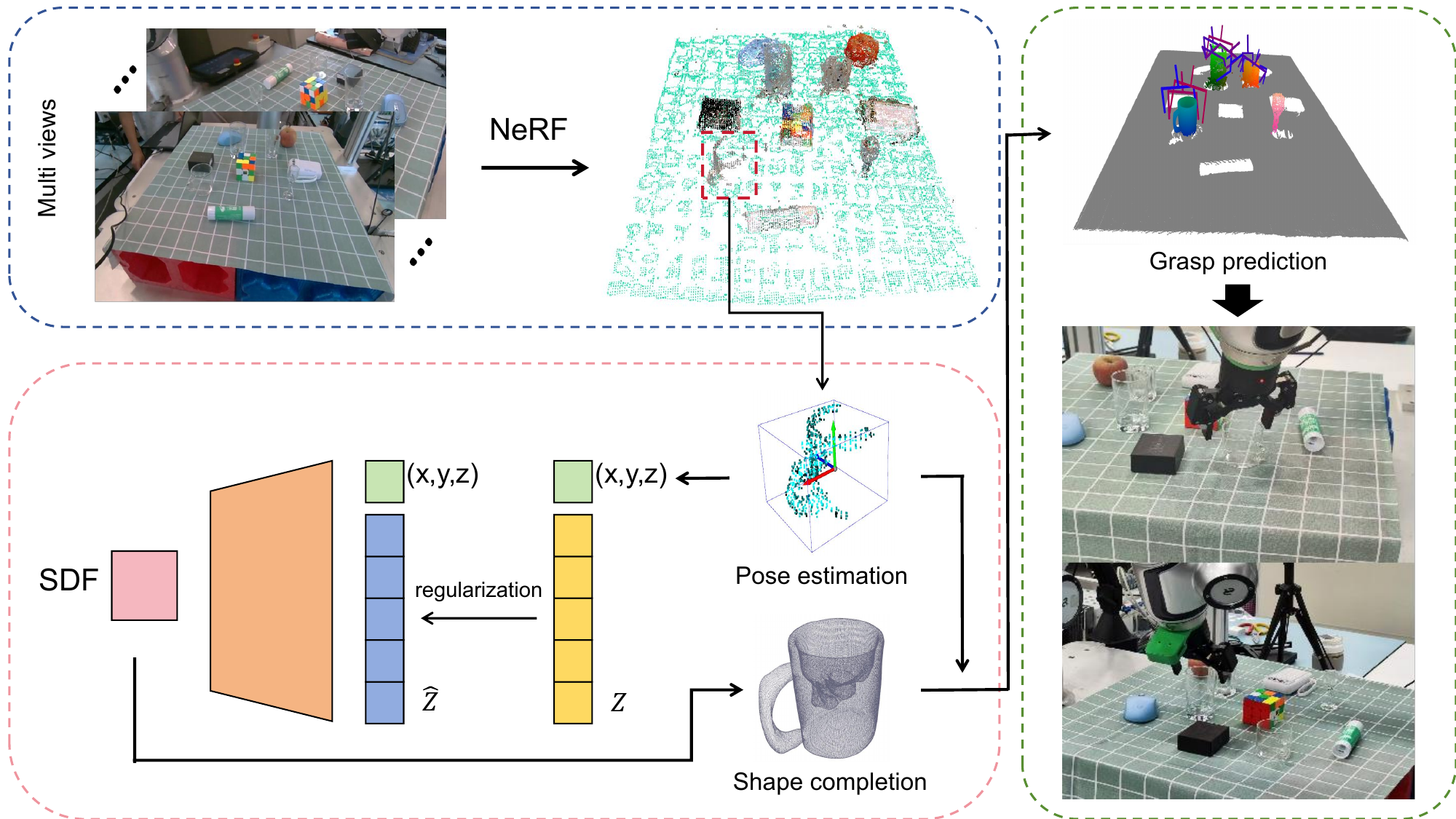}
    \caption{The workflow of our proposed architecture. Starting with NeRF-based scene reconstruction, pose estimation is applied to incomplete transparent objects in the reconstructed results. Subsequently, shape completion is performed using an auto-decoder pre-trained with shape priors. Finally, scene-level grasp predictions for transparent objects are made, followed by validation through experiments on a real robotic system. }
    \label{Fig:workflow}
\end{figure*}

\subsection{System Overview}

Given a cluttered desktop scene containing transparent objects, our objective is to precisely detect and reliably grasp the transparent objects within the scene. In the proposed architecture, multi-view RGB images of the desktop scene are used as input, and the system generates grasp predictions for the target transparent objects based on the panoramically reconstructed scene. To achieve this, we first employ a NeRF-based neural network for panoramic scene reconstruction, which effectively mitigates the challenges associated with acquiring 3D data of transparent objects using conventional depth cameras. Following this, utilizing the 3D spatial structure of the desktop scene, plane fitting and clustering algorithms are applied to segment the point clouds of the objects within the scene. For sparse and incomplete point clouds of non-revolute symmetric transparent objects, we perform pose estimation by exploiting critical geometric regions of the object. The estimated pose facilitates the normalization of the incomplete point clouds, which are subsequently fed into a pre-trained auto-decoder conditioned on shape priors. This auto-decoder reconstructs a complete, dense point cloud of the object’s surface, thereby enabling 3D shape completion. Finally, using the shape-completed point clouds, we refine the previously reconstructed panoramic scene and perform grasp prediction for the transparent objects. These predictions are then validated through experiments on a real robotic system. For a comprehensive explanation of the architecture, please refer to Fig \ref{Fig:workflow}.

\subsection{Panoramic Scene Reconstruction}
\label{od&pe}

Given the presence of transparent objects in the desktop scene, we adopt a NeRF-based neural network for panoramic scene reconstruction, which allows for the accurate retrieval of 3D data that is otherwise difficult to capture using conventional depth cameras. NeRF’s ability to model continuous volumetric opacity enables effective reconstruction of transparent surfaces, whereas traditional depth cameras, relying on principles such as structured light or time-of-flight, struggle to acquire accurate 3D information from transparent objects due to their inability to reflect or scatter light in a manner detectable by these optical sensors.

\textbf{Preliminary: Neural Radiance Fields (NeRFs).}
Neural Radiance Fields (NeRF)\cite{mildenhall2021nerf} represent scenes as a continuous 5D function using a fully connected neural network. This function takes as input a 3D spatial location $x = (x, y, z)$ and a viewing direction $d = (\theta, \phi)$, and outputs the RGB color $c = (r, g, b)$ and volumetric density $\sigma$ at that location. The scene is thus modeled as $F_\Theta(x, d) = (c, \sigma)$, where $\Theta$ are the network parameters.

To render a scene, NeRF calculates the color $C(r)$ along a camera ray $r(t) = o + t d$ by integrating color and density values along the ray:
\begin{equation}
   C(r) = \int_{t_n}^{t_f} T(t) \sigma(r(t)) c(r(t), d) dt, 
    \label{nerf_ray} 
\end{equation}
where $T(t) = \exp \left( - \int_{t_n}^{t} \sigma(r(s)) ds \right)$ represents the transmittance, or the likelihood of the ray traveling without occlusion.

Instant Neural Graphics Primitives (Instant-ngp)\cite{muller2022instant} accelerates NeRF training by using a multi-resolution hash table for encoding 3D coordinates. Rather than traditional positional encoding, Instant-ngp maps 3D points $x$ to feature vectors $f(x)$ via a hash function over multiple levels $l$:
\begin{equation}
   f(x) = \sum_{l=1}^{L} W_l H_l(x), 
    \label{instantngp_hash} 
\end{equation}
where $W_l$ are the weights for level $l$, and $H_l(x)$ is the hash function. This multi-level encoding efficiently captures both fine and large-scale scene details, enabling real-time performance in training and inference.

Instant-ngp’s speed is its key advantage, as it dramatically reduces the time required to learn scene representations. Given our need for fast processing in robotic grasping tasks, Instant-ngp is an ideal choice for reconstructing scenes quickly and accurately, especially in static desktop scenes containing transparent objects.

\textbf{Panoramic Reconstruction of the Desktop Scene.}
We acquire multi-view RGB images of the desktop scene containing transparent objects as input and generate the panoramic reconstruction point cloud of the scene using a neural network based on instant-ngp, as illustrated in Fig \ref{Fig:workflow}. Since NeRF represents the scene as a continuous 5D function, the volumetric density $\sigma$ at each point in the output can encode the transparency of that point, where $\sigma = 0$ denotes complete transparency (typically corresponding to air), and $\sigma = 1$ indicates full opacity (usually corresponding to the surface of opaque objects). For spatial points corresponding to transparent objects, $\sigma$ values are generally non-zero but significantly less than 1, enabling the extraction of spatial points for transparent objects through thresholding. Due to the varying optical properties of different transparent objects, the quality of the 3D information reconstructed by the instant-ngp-based neural network exhibits variation. As shown in Fig \ref{Fig:workflow}, the reconstructed point clouds for certain transparent objects may appear incomplete or sparse.

\subsection{Object Segmentation and Pose Estimation}
\label{sr}

Through panoramic reconstruction, we acquire a point cloud representing the desktop scene, including 3D information of both the surface and the objects present. Object segmentation is employed to distinguish between the point clouds of different objects, a critical step for accurate grasp prediction of transparent objects. Furthermore, for certain non-revolute symmetric transparent objects, pose estimation is conducted to support subsequent point cloud refinement and processing.

\textbf{Object Segmentation.}
In general, a desktop scene can be conceptualized as a composition of the desktop surface and the objects placed upon it, where the spatial arrangement of the objects is closely correlated with the desktop plane. Consequently, we first apply RANSAC plane fitting to extract the fitted plane of the desktop, denoted as $\mathcal{P}_{desk}$, from the panoramically reconstructed point cloud of the scene.

In our scene, considering real-world conditions, it is unrealistic for objects on the desktop to be suspended in mid-air. Therefore, a reasonable assumption can be made: the point clouds of objects are adjacent to the desktop plane $\mathcal{P}_{desk}$, and the surface shapes represented by the object point clouds exhibit continuity (despite the incompleteness of some point clouds, the valid regions generally remain continuous). Based on this assumption, the inherent continuity of object point clouds justifies the use of region growing clustering for object segmentation. Furthermore, the spatial relationship between the objects and $\mathcal{P}_{desk}$ facilitates filtering of the clustering results to eliminate non-object noise, while individual objects are distinguished through their geometric features.

\textbf{Pose Estimation.}
In our scene, the majority of transparent objects exhibit rotational symmetry, with their poses defined by the axis of rotational symmetry (typically perpendicular to the desktop), facilitating straightforward pose estimation. However, certain non-revolute symmetric transparent objects, such as mugs, necessitate specialized pose estimation techniques to support the normalization required for subsequent shape completion processes.

\begin{figure}[!h]
    \centering
    \includegraphics[width=7cm]{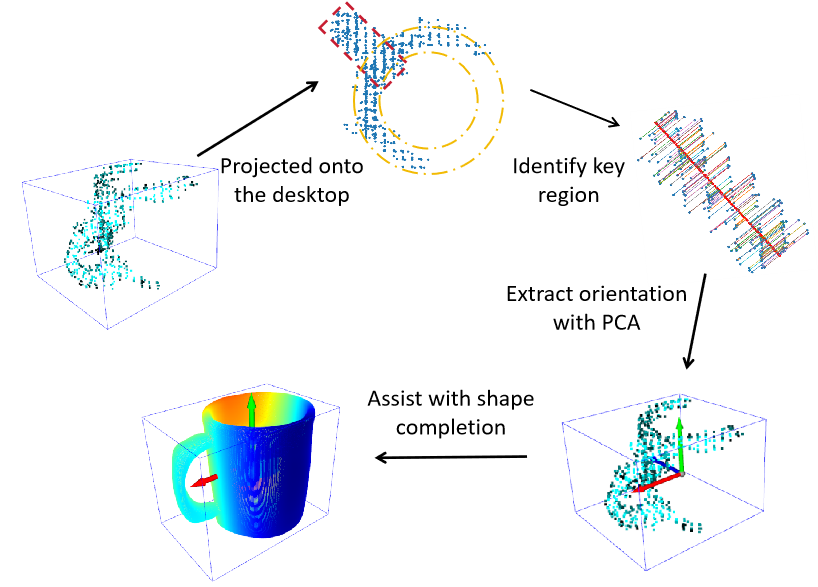}
    \caption{Pose estimation method for non-revolute symmetric objects, with a primary focus on identifying geometric key regions and extracting their orientation. }
    \label{Fig:non_revolute}
\end{figure}

The surface geometry of non-revolute symmetric objects typically exhibits prominent geometric features, such as protrusions relative to the main body, which serve as valuable cues for pose estimation. Taking the mug as an example, as illustrated in Fig \ref{Fig:non_revolute}, we project the point cloud onto the desktop plane $\mathcal{P}_{desk}$. It can be observed that the 2D projection consists of a primary ring (representing the body of the mug) and a protrusion (representing the mug handle), where the orientation of this protrusion directly informs the overall pose of the mug. We define such geometrically salient features, which determine an object's orientation, as key regions. The orientation of these key regions can be leveraged to facilitate pose estimation. In the case of the mug, the orientation of the protrusion key region can be determined using PCA (Principal Component Analysis), representing the orientation of the mug handle. This approach, grounded in geometric key regions, provides an effective solution for pose estimation of non-revolute symmetric objects.

\subsection{Shape Completion}
\label{sr}

In the panoramically reconstructed point cloud of the desktop scene, certain transparent objects (such as the mug) present sparse and incomplete point clouds, which hinder accurate grasp prediction. To address this, we pre-train an auto-decoder that leverages shape priors to achieve shape completion from incomplete surface point clouds.

\textbf{Auto-decoder Framework.}
The auto-decoder functions as an up-sampling network, making it well-suited for reconstructing complete surface geometries from observed incomplete surfaces. Our pre-trained auto-decoder architecture, inspired by DeepSDF\cite{park2019deepsdf}, takes latent codes as input and generates a complete 3D surface point cloud for the target transparent objects. The training process of the auto-decoder leverages shape priors derived from a large set of objects similar to those targeted for shape completion. The latent code acts as an implicit representation of shape information, encoding the spatial positions of numerous points and their corresponding SDF (Signed Distance Function) values. SDF values describe the minimum distance between spatial points and the object surface, with $SDF(\cdot)=0$ serving as an implicit representation of the object's surface.

\textbf{Latent Code Construction.}
Given that the shape priors utilized during the training of the auto-decoder are in standardized size and pose, it is essential to perform normalization on the point cloud to align it with these priors before constructing the latent code. For instance, in the case of the mug, pose normalization can be straightforwardly achieved using the previously estimated pose. Regarding size normalization, due to the incomplete nature of the point cloud, we select a reliable geometric parameter—the mug’s height—as a reference. Specifically, the height of the mug is defined as the vertical distance from the farthest point $p_{max}$ in the point cloud to the desktop plane $\mathcal{P}_{desk}$. Denoting this distance as $d_{max}$ and the standard height of the mug as $h_{mug}$, the normalization scaling factor $\alpha$ is computed as $\alpha = \frac{h_{mug}}{d_{max}}$. This scaling factor is then uniformly applied to the entire point cloud to ensure dimensional consistency with the shape priors.
Following the normalization of the incomplete point cloud, SDF values are assigned to each point (commonly set to 0, as the majority of the points lie near the object's surface). This process finalizes the construction of the latent code for shape completion.

\subsection{Grasp Prediction}
Substitute the incomplete object point clouds within the panoramically reconstructed scene with the shape-completed point clouds. We utilize a model derived from GraspNet-1billion \cite{fang2020graspnet} to predict grasp poses for the transparent objects in the scene and subsequently implement these plausible grasp predictions on a physical robotic system. GraspNet-1billion is specifically designed for scene-level grasp prediction, with models well-suited for environments where objects are positioned in a cluttered manner, making it an appropriate choice for our task.
To facilitate deployment on a real robotic system, we normalize the panoramic reconstruction scene point cloud to align it with the depth camera-captured point cloud before conducting grasp prediction.

\section{Experiments}

Experiments are conducted in a vision-based robotic manipulation system which consists of a UR3e robot manipulator, a two-finger parallel gripper (Robotiq), a 3D camera (Mech-Eye LOG M Industrial), and a PC with Ubuntu 20.04 LTS (CPU: Intel Core i9-10940X, RAM: 16GB, GPU: NVIDIA GeForce RTX 3090).

Our proposed architecture needs to validate the following aspects: (1) the effectiveness of our panoramic scene reconstruction method in acquiring 3D information of transparent objects; (2) the enhancement of grasp prediction through our shape completion method; (3) the overall improvement in grasping transparent objects within the scene by our architecture. Therefore, the experimental section is divided into three modules. The desktop scene used in the experiments is custom-built by us, containing several transparent objects of various shapes and some non-transparent objects serving as background, with the arrangement of objects being random.

\subsection{Panoramic Scene Reconstruction}
\begin{figure} [!h]
  \centering
  \includegraphics[width=8cm]{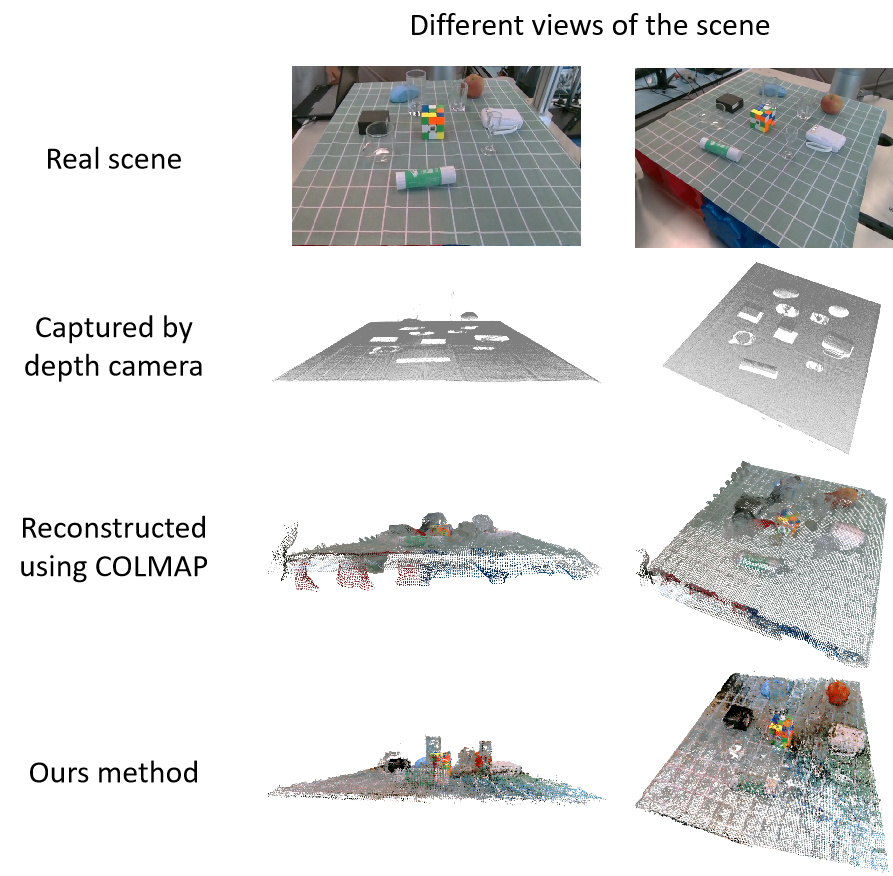}
      
  \caption{Scene point clouds obtained by various methods. Transparent object 3D data from the depth camera and COLMAP\cite{schoenberger2016sfm, schoenberger2016mvs} is severely missing and distorted, making grasp prediction impossible. In contrast, our method yields reliable 3D information for transparent objects.}
  \label{Fig:scene_reconstruction}

\end{figure}

For the custom-built desktop scene, we obtained the scene point clouds using a depth camera fixed above the scene, COLMAP\cite{schoenberger2016sfm, schoenberger2016mvs}, and our instant-ngp-based method, as illustrated in Fig \ref{Fig:scene_reconstruction}.

Conventional depth cameras are inherently limited in their ability to detect light reflection from transparent objects, resulting in a near-complete absence of 3D information for these objects in the point cloud, with their positions represented by voids and noise. COLMAP, which leverages Structure-from-Motion (SfM) and Multi-View Stereo (MVS) techniques, relies on feature point matching across multiple RGB views. However, the absence of texture and color on transparent objects significantly interferes with this process. Consequently, in the reconstructed point clouds, transparent objects exhibit substantial blending with their surroundings, making object segmentation difficult and leading to severe inaccuracies in 3D information, rendering it unsuitable for grasp prediction. In contrast, our method enables the continuous modeling of opacity in the scene, facilitating more accurate and comprehensive panoramic reconstructions of the desktop environment. For transparent objects, our approach ensures high density and accuracy of the reconstructed 3D data. Even when point cloud deficiencies occur for certain objects, the available 3D information is sufficient to support reliable shape completion.

\subsection{Shape Completion}

\begin{figure} [!h]
  \centering
  \includegraphics[width=8cm]{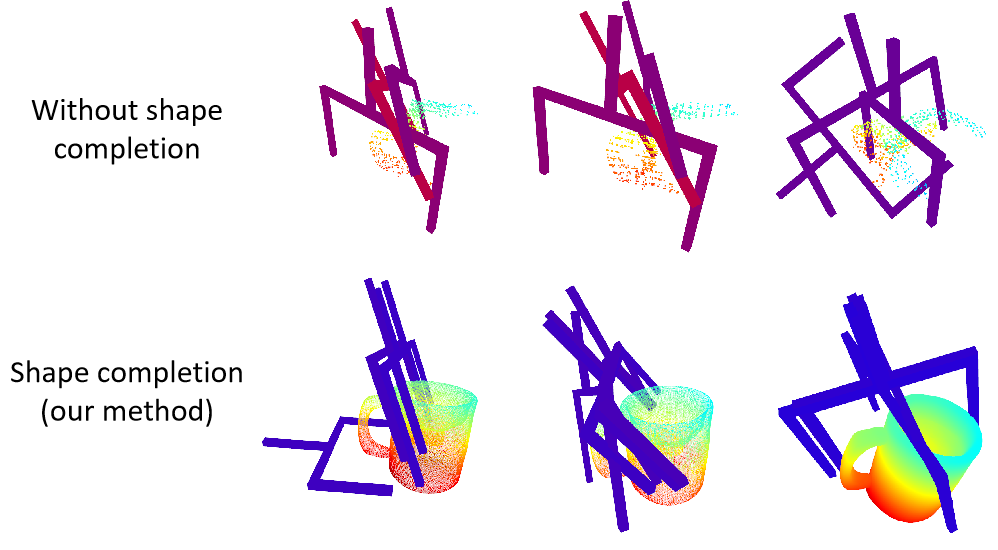}
      
  \caption{Comparison of grasp predictions before and after shape completion, with higher prediction quality following completion.}
  \label{Fig:shape_completion}

\end{figure}

In the panoramic reconstruction of the scene point cloud, the non-revolution-symmetric transparent mug, exhibiting noticeable deficiencies in its point cloud, is selected as the target for shape completion. To assess the effectiveness of our shape completion approach for transparent object grasping tasks, we evaluate the grasp predictions for the mug both before and after the shape completion process.
The reconstruction results are shown in Fig. \ref{Fig:shape_completion}.

\begin{table}[h]
\renewcommand\arraystretch{1.0}
\centering
\caption{Scores of grasp predictions generated by the GraspNet-1billion model\cite{fang2020graspnet} affected by shape completion.}
\label{tab:scores}
\resizebox{1.0\linewidth}{!}{
\begin{tabular}{cccccc}
\toprule
\begin{tabular}[c]{@{}c@{}}Shape completion\\or not?\end{tabular} &
\begin{tabular}[c]{@{}c@{}}Maximum\end{tabular} &
\begin{tabular}[c]{@{}c@{}}Average\\(top 5)\end{tabular} &
\begin{tabular}[c]{@{}c@{}}Average\\(top 10)\end{tabular} &
\begin{tabular}[c]{@{}c@{}}Variance\\(top 5)\end{tabular} &
\begin{tabular}[c]{@{}c@{}}Variance\\(top 10)\end{tabular}\\ 
\midrule
  No&0.7259&0.4799&0.4273&0.0196&0.0118\\
  Yes(ours)&0.9327&0.8132&0.7485&0.0078&0.0087\\
\bottomrule
\end{tabular}
}
\end{table}

Grasp predictions are generated using a model based on GraspNet-1billion\cite{fang2020graspnet}, which predicts 6-DoF grasp poses around the target object, assigning a score to each prediction to evaluate its quality. From a qualitative perspective, as illustrated in Fig \ref{Fig:shape_completion}, the incomplete point cloud of the mug results in significant prediction errors, such as gripper trajectories penetrating the object's surface and severe limitations in the number of feasible grasp poses. The former issue stems from the disruption of collision detection due to missing surface geometry, while the latter arises from the model's inability to generate grasp poses without sufficient point cloud data. These deficiencies are substantially mitigated after shape completion. Quantitatively, as shown in Table \ref{tab:scores}, both the maximum score and average scores significantly increase post-completion, indicating an overall improvement in grasp prediction quality. Furthermore, the variance in prediction scores decreases, reflecting greater consistency and reduced likelihood of low-quality predictions. In conclusion, shape completion substantially improves the quality, stability, and physical accuracy of grasp predictions.

\subsection{Scene-level Transparent Object Grasping}

\begin{figure} [!h]
  \centering
  \includegraphics[width=8.5cm]{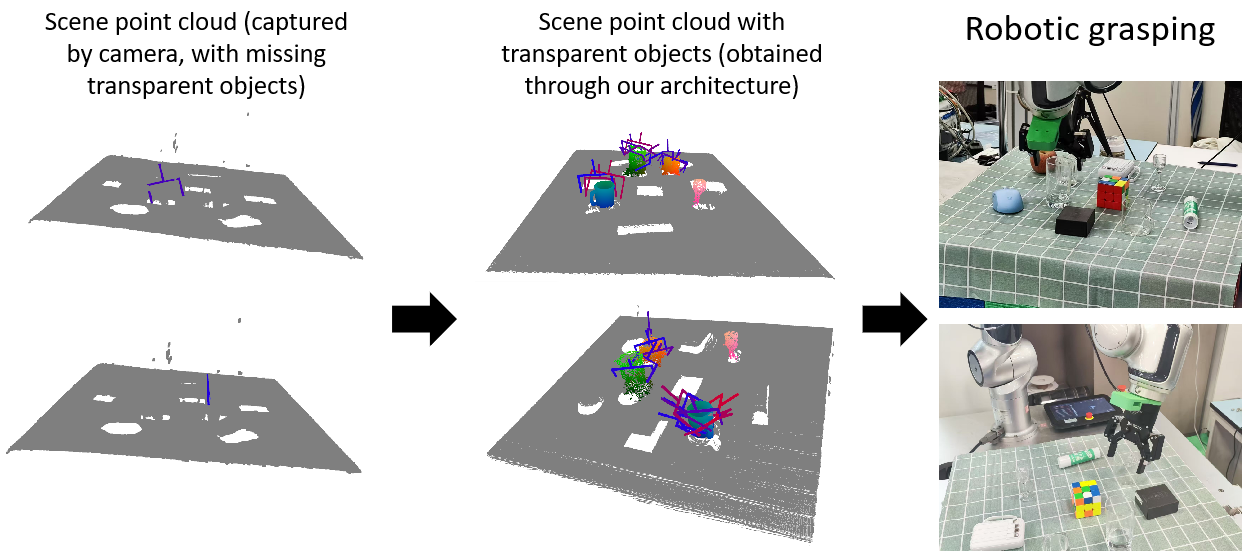}
  \caption{Scene-level grasp prediction for transparent objects and deployment on a real robot.}
  \label{Fig:grasp_prediction}
\end{figure}

By leveraging panoramic reconstruction and shape completion, we obtain reliable 3D data of transparent objects, which necessitates validation through scene-level grasp prediction and real-world robotic experiments to confirm the efficacy of our architecture. As illustrated in Fig \ref{Fig:grasp_prediction}, the point cloud captured by the depth camera demonstrates significant 3D information loss for transparent objects, rendering effective grasp prediction nearly impossible. However, after processing through our architecture, accurate 3D reconstructions of transparent objects are achieved, enabling feasible and effective grasp predictions for the objects, most of which can be successfully executed in a physical robotic system. (Since grasp predictions generated by the GraspNet-1billion model depend on object size, the gripper width must be adjusted accordingly. Therefore, in Fig \ref{Fig:grasp_prediction}, no grasp predictions are displayed around the smaller wine glass, in contrast to the larger cups.)

\begin{table}[h]
\renewcommand\arraystretch{1.0}
\centering
\caption{Execution success rate of grasp predictions for transparent objects within the scene (\%).}
\label{tab:success_rate}
\resizebox{1.0\linewidth}{!}{
\begin{tabular}{ccccc}
\toprule
\begin{tabular}[c]{@{}c@{}}\textbf{Objects}\end{tabular} &
\begin{tabular}[c]{@{}c@{}}Cylindrical cup\end{tabular} &
\begin{tabular}[c]{@{}c@{}}Prismatic cup\end{tabular} &
\begin{tabular}[c]{@{}c@{}}Mug\\(incomplete)\end{tabular} &
\begin{tabular}[c]{@{}c@{}}Mug\\(complete)\end{tabular} \\ 
\midrule
  success rate&\textbf{90.77}&\textbf{87.72}&27.27&\textbf{83.78}\\
\bottomrule
\end{tabular}
}
\end{table}

We analyzed the execution success rates of the grasp predictions generated for each transparent object within the scene. A successful execution is defined as a grasp prediction that can be realistically performed in the physical world. As presented in Table \ref{tab:success_rate}, all transparent objects demonstrated high execution success rates. Notably, the substantial improvement in the success rate for the mug following shape completion underscores the critical role of shape completion in enhancing grasp prediction accuracy.

\section{Conclusion}
In this paper, we present a NeRF-based architecture for transparent object grasping, augmented with shape priors. The proposed architecture first employs a NeRF-based neural network to perform panoramic scene reconstruction, effectively overcoming the inherent challenges in acquiring 3D information for transparent objects. Subsequently, object segmentation and pose estimation are carried out using non-learning, geometry-driven methods. A pre-trained auto-decoder, leveraging shape priors, is then utilized to complete the shape of partially reconstructed transparent objects. Finally, grasp predictions are generated on the enhanced scene point cloud, now enriched with the recovered 3D information of transparent objects. The validity of our architecture is substantiated through several key findings: (1) the NeRF-based scene reconstruction method reliably acquires 3D information for transparent objects; (2) the shape-prior-guided completion method significantly improves grasp prediction in terms of quality, stability, and real-world applicability; and (3) our architecture consistently achieves high grasp execution success rates for transparent objects within various scenes.

\addtolength{\textheight}{0cm}   








\bibliographystyle{IEEEtran}
\bibliography{IEEEexample}

\end{document}